\documentclass[runningheads]{llncs}
\usepackage{graphicx}

\usepackage{multirow}

\usepackage{subfigure}
\usepackage{wrapfig}
\usepackage{diagbox}
\usepackage[linesnumbered,ruled,vlined]{algorithm2e}
\usepackage{algpseudocode}

\usepackage{graphicx}
\usepackage{amsmath}
\usepackage{amssymb}
\usepackage{booktabs}

\usepackage{multirow}
\usepackage{multicol}
\usepackage{hyperref}
\hypersetup{hypertex=true,
colorlinks=true,
linkcolor=blue,
anchorcolor=blue,
citecolor=blue}

\begin{document}
\title{WsiCaption: Multiple Instance Generation of Pathology Reports for Gigapixel Whole-Slide Images}

\author{Pingyi Chen\inst{1,2,3}
\and
Honglin Li \inst{1,2,3}\and
Chenglu Zhu\inst{2,3}\and
Sunyi Zheng\inst{2,3}\and
Zhongyi Shui\inst{1,2,3}\and
Lin Yang\inst{2,3}}

\authorrunning{P. Chen et al.}

\institute{College of Computer Science and Technology, Zhejiang University \and
Research Center for Industries of the Future, Westlake University\and School of Engineering, Westlake University.
\\
\email{\{chenpingyi,yanglin\}@westlake.edu.cn}}

\titlerunning{WsiCaption}
%

%

\maketitle              
\begin{abstract}
Whole slide images are the foundation of digital pathology for the diagnosis and treatment of carcinomas. Writing pathology reports is laborious and error-prone for inexperienced pathologists. To reduce the workload and improve clinical automation, we investigate how to generate pathology reports given whole slide images. On the data end, we curated the largest WSI-text dataset (PathText). In specific, we collected nearly 10000 high-quality WSI-text pairs for visual-language models by recognizing and cleaning pathology reports which narrate diagnostic slides in TCGA. On the model end, we propose the multiple instance generative model (MI-Gen) which can produce pathology reports for gigapixel WSIs. We benchmark our model on the largest subset of PathText. Experimental results show our model can generate pathology reports which contain multiple clinical clues and achieve competitive performance on certain slide-level tasks. We observe that simple semantic extraction from the pathology reports can achieve the best performance (0.838 of F1 score) on BRCA subtyping surpassing previous state-of-the-art approaches. Our collected dataset and related code are available at \href{https://github.com/cpystan/Wsi-Caption}{https://github.com/cpystan/Wsi-Caption}.

\end{abstract}

\section{Introduction}
\label{sec:intro}

Whole-slide images (WSI) based diagnostic pathology is the foundation and gold standard for the diagnosis of carcinoma. Due to the enormous size and large amount of heterogeneous information that exists in WSIs, the reading and interpretation of the slide usually necessitates specialized pathologists. Recently, with the advancement of computer-aided methods of WSIs, computational pathology has achieved remarkable success and, assisted with deep learning, some
can even outperform experienced pathologists in certain tasks \cite{shao2021transmil,zhang2022dtfd,chan2019histosegnet,psm}. These advanced methods have largely improved the automation of the pathological reading workflow, especially for those less-experienced pathologists in rural areas \cite{zhang2019pathologist}.

In spite of the "clinical-grade" performance of these computational pathology approaches, pathologists still need to organize the findings and write textual reports for each slide. Hundreds to thousands of WSIs need to be summarized in text in the pathology departments every day \cite{farahani2015whole}. The automation of diagnostic reports can largely reduce the workload of pathologists. Furthermore, the content of pathology reports usually includes abundant diagnostic results \cite{buckley2012feasibility}. Therefore, it motivates us to take a step forward to achieve the automatic generation of pathology reports. On the data end, the great advancement of computational pathology in the past years owes very much to the emergence of publicly available pathology datasets. Some researchers resort to books, articles, and webs \cite{gamper2021bookandarticle,lu2023mizero,huang2023plip} to obtain large-scale image-text pairs. However, their collected images are small patches and the corresponding texts are also limited to patch-level descriptions. Therefore, collecting high-quality WSI-text pairs is worth exploring and can boost the development of visual-language models in computational pathology.

We notice that TCGA includes scanning copies of pathology reports in the format of PDF\footnote{\href{https://portal.gdc.cancer.gov/}{https://portal.gdc.cancer.gov/}}. But they are too long with redundant information and present in a complex structure. Therefore, we propose a pipeline to extract and clean pathological texts from TCGA, which can convert complex PDF files to concise WSI-text pairs with the assistance of large language models (LLM).


There are still challenges to achieving slide-level generation on the model end. Recent years have witnessed the boom of visual-language models in image captioning \cite{vinyals2015showandtell,xu2015showattendandtell,rennie2017selfcritical}. And in the medical area, the generation of radiology reports has been explored by several works \cite{jing-etal-2018-automatic,chen-emnlp-2020-r2gen,liu2019clinically}. However, it is unaffordable to directly process the WSIs with more than 10 gigapixels unless WSIs are resized or disentangled sacrificing much fine-grained information.  To deal with this problem, we introduce a \textbf{M}ultiple \textbf{I}nstance \textbf{Gen}eration (MI-Gen) framework to achieve WSI-based report generation. It contains a visual extractor which encodes the non-overlap patches with the sliding window and a sequence-to-sequence generator to produce pathology texts.
In this work, our contributions can be concluded as follows:

\begin{enumerate}    
    \item We propose a pipeline to curate high-quality WSI-text pairs from TCGA. The dataset (PathText) contains about ten thousand pairs which will be publicly accessible. It can potentially promote the development of visual-language models in pathology.
    
    \item We design a multiple instance generation framework (MI-Gen) and benchmark our approach with various backbones in the dataset.

    \item Our model can achieve a promising performance on certain slide-level tasks like tumor subtyping, surpassing previous state-of-the-art MIL methods.

\end{enumerate}

\section{Multiple Instance Generation: Problem Formulation}
In the multiple instance generation framework,  the huge image can be seen as a bag which contains a group of instances which are non-overlap patches with a much smaller resolution. Denote the cohort of instances as $Bag(X_i) = \{x_i^j\}_{j=1}^{M_i}$ where $X_i$ is the $i$-th sample in the dataset and $M_i$ is the sequence length which is determined by the patch size. Usually, when the patch size is 256, the sequence length $M$ can be larger than ten thousand. A visual extractor $h$ is adopted to extract image embeddings from the patches, denoted as $h(\{x_i^j\})\in \mathbb{R}^{M_i\times l}$ where $l$ is the embedding size.

A sequence-to-sequence model is incorporated to generate the target sequence $Y_i=\{y_i^j\}_{j=1}^{N_i}$ where $N_i$ is the ground truth sequence length. Like previous text generative models, the encoder-decoder parameterized by $\phi$ is trained using language model objective which maximizes the sum of conditional possibilities of individual words in the sequence. Therefore, the negative log-likelihood (NLL) loss can be calculated as:

\begin{equation}
    L= -\sum_i\sum_{t=1}^N logp_{\theta}(y_t|h(\{x_i^j\}),\{y_i^j\}_{j<t};\phi).
\label{equation:loss}
\end{equation}

The probability of $j$-th word is calculated based on the patch embeddings and the previous sequence $\{y_i^j\}_{j<t}$.

\section{Method}


\begin{figure}[t]
  \centering
   \includegraphics[width=1.0\linewidth]{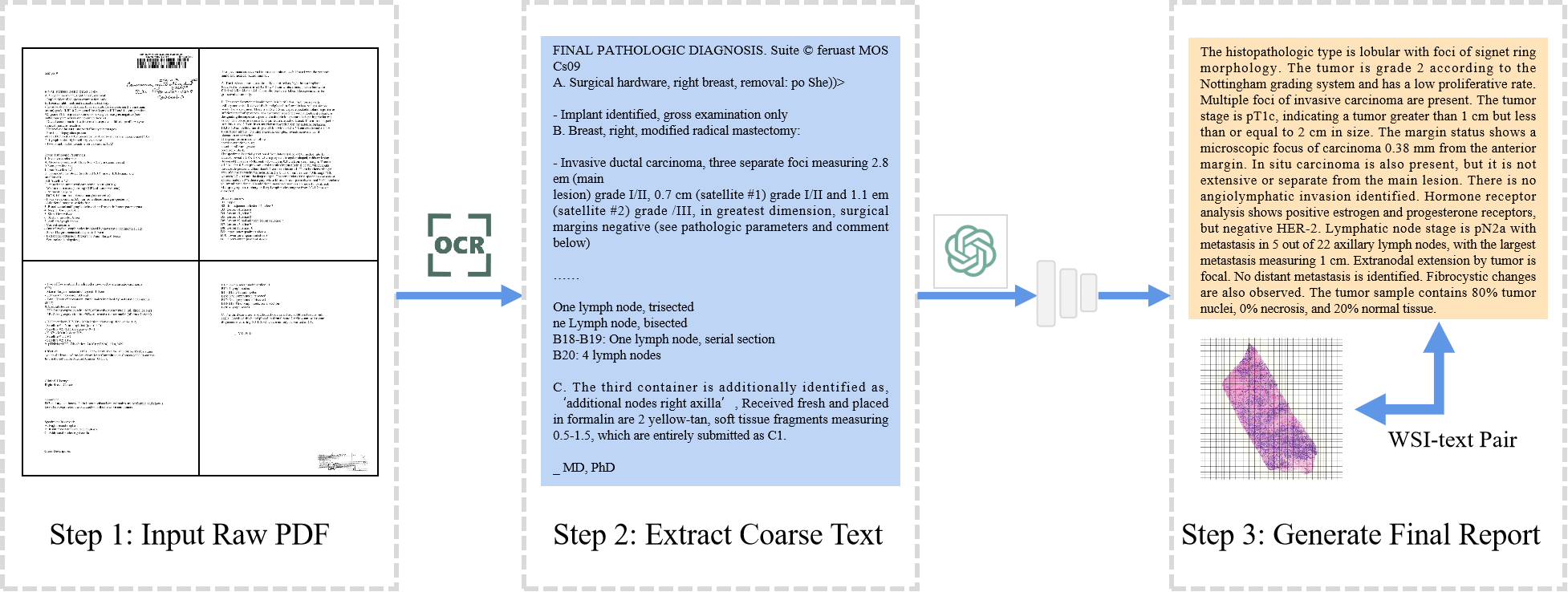}

   \caption{The pipeline of extracting WSI-text pairs from TCGA. }
   \label{fig:dpt}
\end{figure}

\subsection{PathText Construction}
The first step in constructing PathText is to find out the diagnostic slides and their corresponding pathology reports in the TCGA, as shown in Fig. \ref{fig:dpt}. The diagnostic slides from TCGA cover diverse disease types originating at different primary sites. The pathology reports in the format of PDF usually contain multiple pages. We use OCR methods to transform the files into text \cite{smith2007ocr}. However, the text is still noisy because the report itself contains redundant information, and OCR sometimes generates garbled code. Therefore, we apply LLMs \cite{brown2020gpt3} to filter and summarize the report with the prompt of \textit{"Please summarize the following pathology report:"}. And to eliminate the influence of hallucination of LLMs, we manually annotate 88 pairs and train a classifier to remove those reports that are flawed.

PathText contains 9009 WSI-text pairs in total. Our collected text is distilled from clinical pathology reports owning well-aligned correspondence with WSIs and abundant pathological content. Our text is much longer than ARCH \cite{gamper2021bookandarticle} (patch-level descriptions collected from books and articles). More details about PathText are demonstrated in the \textbf{Supplementary Materials}.

\begin{figure*}[t]
  \centering
   \includegraphics[width=1.0\linewidth]{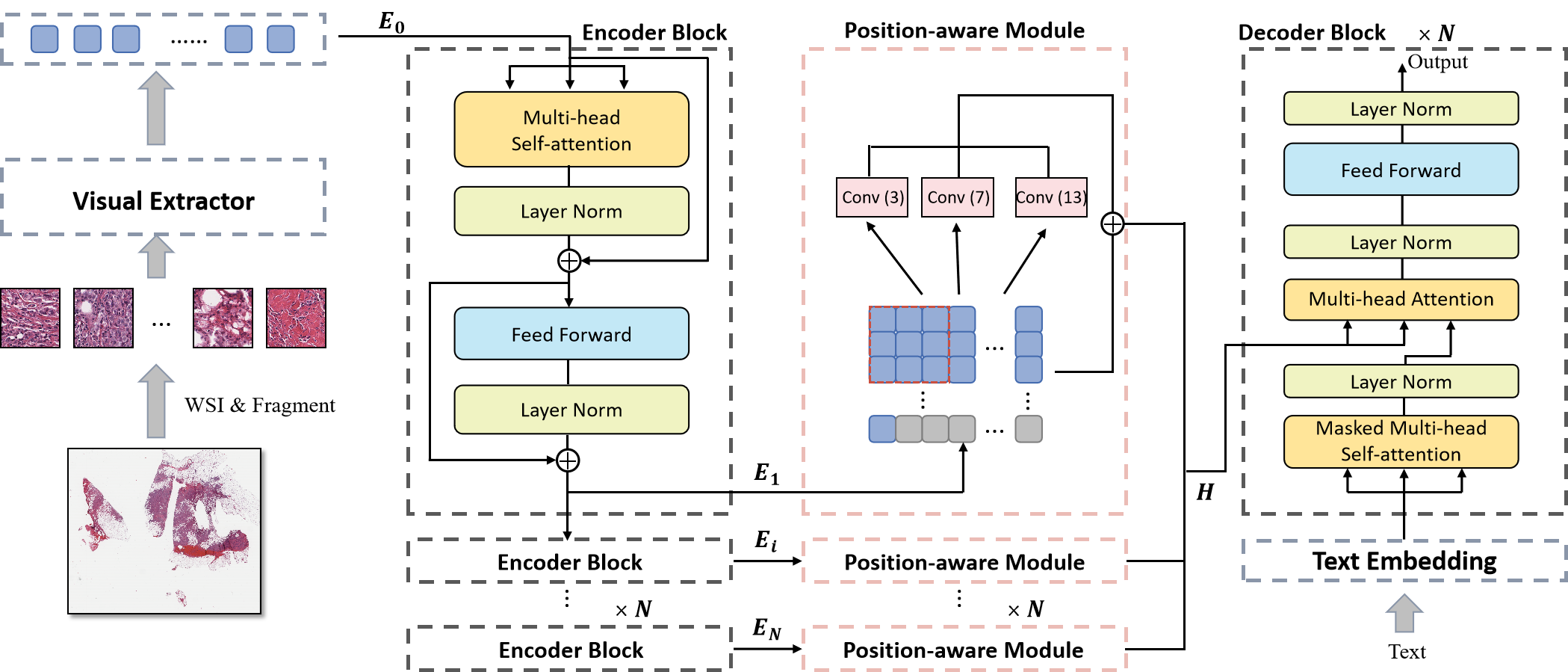}

   \caption{The framework of our proposed model is comprised of a visual extractor and an encoder-decoder. }
   \label{fig:framework}
\end{figure*}

\subsection{WSI-text Generation}
As illustrated in Fig. \ref{fig:framework}, our proposed generative model consists of two parts: the visual extractor, and the encoder-decoder. We incorporate the hierarchical position-aware module into Transformer encoder layers so as to strengthen the model awareness of spatial information in WSIs. The details are described below.

\subsubsection{Visual Extractor.}
Given a whole-slide image $X_i$ with huge resolution, it is comprised of small patches $ \{x_i^j\}_{j=1}^{M_i}$. It is unable to accumulate gradients in the visual extractor due to the large size of WSIs. Therefore, we apply pre-trained neural networks to extract features from these patches. The visual extractor is non-trainable during training. The initial image embeddings are denoted as $\textit{\textbf{E}}_0$ which are fed to the subsequent modules.

\subsubsection{Encoder-Dcoder.}
Transformer has shown remarkable success in the past few years due to its strong capability of modeling the long sequence and interaction among individual tokens with self-attention. Therefore, we adopt Transformer as our backbone. The Transformer encoder consists of $N$ transformer layers. The embeddings are processed by the transformer layers iteratively as:

\begin{equation}
    \textit{\textbf{E}}_i = f_i(\textit{\textbf{E}}_{i-1})
\end{equation}

where $f_i(\cdot)$ refers to the $i$-th transformer layer. Whereas, we propose hierarchical position-aware modules to aggregate the embeddings from different encoder layers. The position-aware modules are inserted after each encoder layer so that more abundant context information is captured. Therefore, the hidden state $\textit{\textbf{H}}$ for decoding can be formulated as:

\begin{equation}
    \textit{\textbf{H}} = \sum_i PAM_i(\textit{\textbf{E}}_i).
\end{equation}

\subsubsection{Position-aware Module.}  It has been confirmed in \cite{shao2021transmil} that convolutional layers improve position information awareness. Inspired by this, we incorporate a hierarchical position-aware module (PAM) into the encoding of image embeddings. Considering the varying sizes of tokens in WSIs, we also conduct padding so that the feature map can be reshaped for fitting convolutional layers. Convolution kernels of various sizes are adopted to capture heterogeneous spatial information. 

In specific, denote the sequence of image embeddings as $\textit{\textbf{E}}_i=\{e_j\}_{j=1}^{M_i}$. Firstly, we pad the sequence until its length becomes a square number $T_i$. Then, the 1-D sequence can be reshaped into 2-D space $\textit{\textbf{E}}_i^{pad} \in\mathbb{R}^{\sqrt{T_i} \times \sqrt{T_i} \times l} $. We adopt several CNNs to process and aggregate the 2-D embeddings:

\begin{equation}
    PAM_i(\textit{\textbf{E}}_i)= \sum Conv_i(\textbf{E}_i^{pad}) +\textbf{E}_i^{pad}.
\end{equation}

Heterogeneous spatial information from the CNNs with varying kernel sizes is summed together. Finally, the hidden state returns to 1-D space as a sequence for decoding. Our PAMs are inserted hierarchically after each encoder block, encoding abundant spatial information at different depth, which improves the awareness of spatial descriptions in the generated reports.

\begin{table*}[t]
\caption{Quantitative results of pathology report generation on PathText (BRCA). Different combinations of visual extractors and encoder-decoders are present for comparison. BLEU-n indicates the BLEU score computed based on n-grams.}
\renewcommand\arraystretch{1.3}
  \centering
  
  \resizebox{\linewidth}{!}
{
    \scriptsize
  \begin{tabular}{l|l|ccccccc}
   \toprule
       Visual Extractor \& Pre-train&Encoder-Decoder &BLEU-1&BLEU-2&BLEU-3&BLEU-4&METEOR&ROUGE& $Fact_{ent}$\\
   \hline
  \multirow{5}{*}{ResNet\&ImageNet}  &CNN-RNN\cite{vinyals2015showandtell} &0.334&0.209&0.122&0.074&0.137&0.248&0.396\ \\
   ~& att-LSTM\cite{xu2015showattendandtell}&0.367&0.234&0.128&0.085&0.151&0.262&0.442\\
  ~& vanilla Transformer \cite{transformer}&0.395 &0.230&0.135&0.089&0.145&0.254&0.453 \\
  ~& Mem-Transformer \cite{chen-emnlp-2020-r2gen} &0.317&0.207&0.136&0.091&0.129&0.270&0.467\\
  \cline{2-9}
  ~& Ours &\textbf{0.403} &\textbf{0.254} &\textbf{0.168}& \textbf{0.117} &\textbf{0.163}&\textbf{0.280}&\textbf{0.498}\\
  \hline

    \multirow{5}{*}{ViT\&ImageNet}  &CNN-RNN\cite{vinyals2015showandtell} &0.328&0.201&0.127&0.082&0.142&0.253&0.380\\
   ~& att-LSTM\cite{xu2015showattendandtell}& 0.341&0.211&0.132&0.083&0.145&0.265&0.425\\
  ~& vanilla Transformer\cite{transformer}&0.346&0.216&0.137&0.091&0.149&0.273&0.443\\
  ~& Mem-Transformer\cite{chen-emnlp-2020-r2gen}&0.332&0.216&0.144&0.100&0.147&0.268&0.449\\
  \cline{2-9}
  ~& Ours&\textbf{0.380} &\textbf{0.252} &\textbf{0.169}& \textbf{0.110} &\textbf{0.157}&\textbf{0.279}&\textbf{0.474}\\
  \hline

    \multirow{5}{*}{ViT\&HIPT} & CNN-RNN\cite{vinyals2015showandtell} &0.342&0.215&0.141&0.084&0.148&0.260&0.403 \\
   ~& att-LSTM\cite{xu2015showattendandtell} &0.372&0.230&0.135&0.090&0.150&0.269&0.466\\
  ~&  vanilla Transformer\cite{transformer} &0.383&0.237&0.151&0.096&0.152&0.264&0.488\\
  ~& Mem-Transformer\cite{chen-emnlp-2020-r2gen}&0.344&0.218&0.150&0.103&0.151&0.268&0.501\\
  \cline{2-9}
  ~& Ours&\textbf{0.446} &\textbf{0.286} &\textbf{0.183}& \textbf{0.120} &\textbf{0.171}&\textbf{0.271}&\textbf{0.532}\\

\bottomrule
  \end{tabular}}

\label{tab:comp}
\end{table*}

\section{Experiments and Results}
\subsection{Implementation Details}
\subsubsection{Datasets.} We train and validate our generative model on the PathText (BRCA) which includes 845 pairs for training, 98 pairs for validating, and 98 pairs for testing. TCGA-BRCA contains 1041 whole slide images with the label of invasive ductal (IDC) or invasive lobular carcinoma (ILC). It also contains the results of Her2 testing.  10$\%$  of TCGA-BRCA  is randomly selected for validation and inference. 
 
\subsubsection{Model Setting.} The number of encoder layers and decoder layers are both 3. For self-attention modules, the number of heads is 4 and the size of embeddings is 512. Three CNNs are adopted in the PAM with the kernel size of 3, 7, and 13 respectively. We use Adam with the learning rate of 1e-4 to optimize the model. The weight decay is 5e-5. We adopt beam search with the size of 3 as the sampling method. Our model is trained on four A100-80G GPUs.

\begin{figure*}[t]
  \centering
   \includegraphics[width=1.0\linewidth]{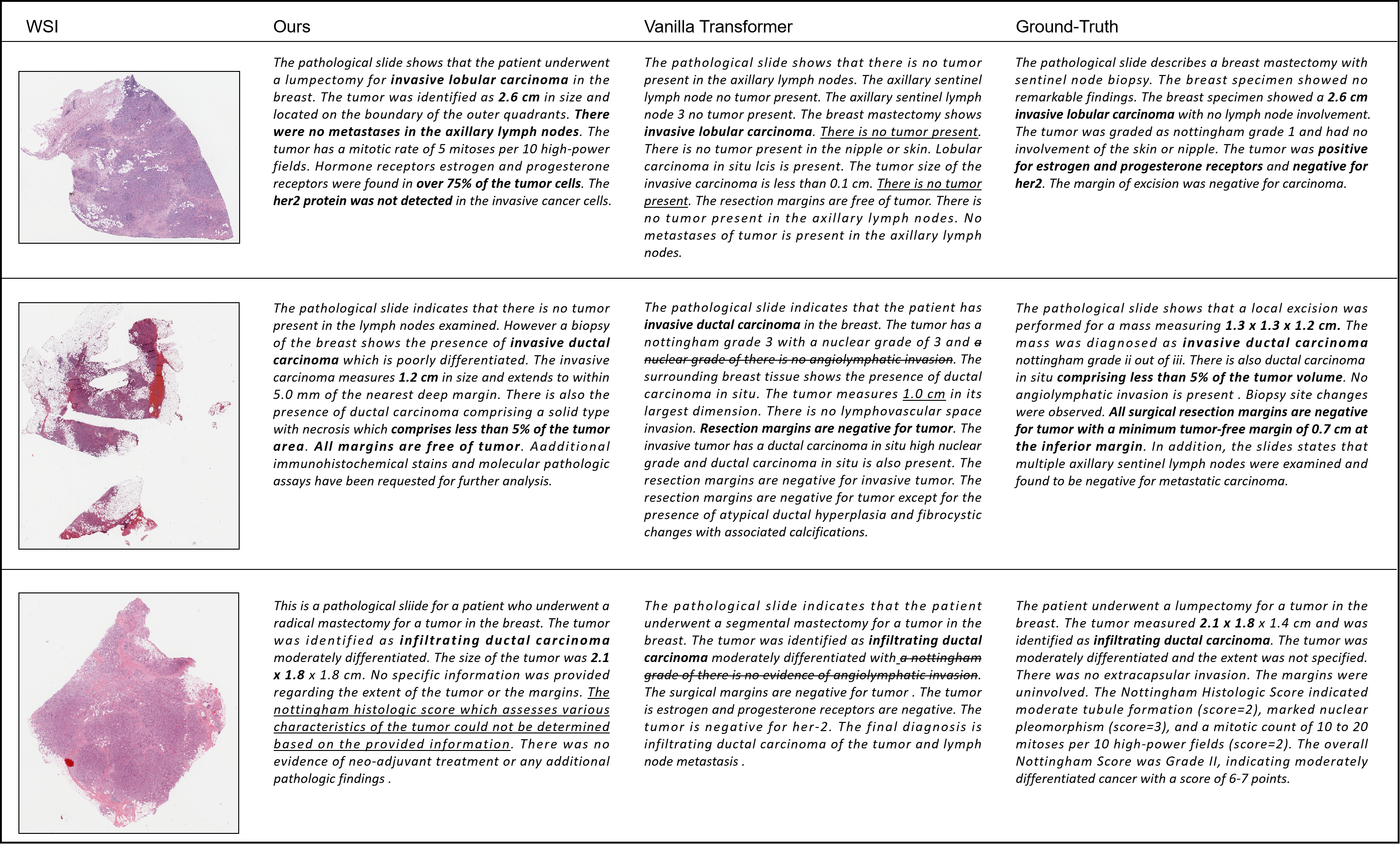}

   \caption{Illustrations of pathology reports from our model, Vanilla Transformer, and ground-truth. The first column shows the thumbnails of the WSIs. The content that is consistent with the ground-truth is highlighted in bold. And the medical terms which are contradictory to ground-truth are underlined. Strikethrough text means there existing grammar errors.}
   \label{fig:vis}
\end{figure*}

\subsection{Baselines}
To benchmark our models, two kinds of visual models are chosen as our visual extractor: ResNet \cite{resnet} which is composed of convolutional layers and ViT \cite{vit} which is based on Transformer. For the pre-training of visual extractors, we explore two strategies: 1) ImageNet (out-of-domain) pre-training on extensive natural images, and 2) hierarchical self-supervised learning with a pyramid transformer (HIPT) \cite{hipt} on TCGA (in-domain). 

We also apply several other popular backbones. There are two backbones which decode by LSTMs: CNN-RNN \cite{vinyals2015showandtell} and att-LSTM \cite{xu2015showattendandtell}. We set the hidden size of these two LSTM-based models as 512 and the number of layers is 3. We also re-implemented two state-of-the-art Transformer-based models. The first is Vanilla Transformer which can be seen as the ablated version of our method with the same structure except for the position-aware module. The other is Mem-Transformer \cite{chen-emnlp-2020-r2gen} which is specially designed for medical report generation by incorporating a memory mechanism in the decoder.

\begin{table}[t]
    \caption{
    The performance of slide-level tasks. Different MIL approaches are included for comparison. P, R, and F1 indicate precision, recall, and F1 score respectively.}
  \begin{center}  
  \resizebox{0.7\linewidth}{!}{
  \begin{tabular}{c|ccc|ccc}
  

   \hline
     \multirow{2}{*}{\diagbox[]{Method}{Tasks}} & \multicolumn{3}{c|}{{Her2-prediction}} & \multicolumn{3}{c}{{Subtyping }}   \\
     \cline{2-7}
    &P&R & F1 & P & R & F1  \\
  \hline
  
  Max-pooling & 0.545 & 0.601 & 0.570&  0.612 & 0.662&0.644\\

  AB-MIL \cite{ilse2018attention} &  0.622 & 0.670 & 0.645 & 0.730&0.823&0.771\\
  DS-MIL \cite{li2021dual} &  0.629& 0.684 &  0.652 & 0.727&0.821&0.775 \\
CLAM-SB \cite{lu2021clam} &  0.605& 0.668 &  0.632& 0.781& 0.879&0.823\\
TransMIL \cite{shao2021transmil} & 0.649& 0.701 &  0.670&0.782&0.834& 0.806\\

  \hline


Semantic Extraction&\textbf{0.655}  & \textbf{0.710} & \textbf{0.678}& \textbf{0.805}& 0.865&\textbf{0.838}\\
 \hline
  \end{tabular}   }

  \end{center}   

    \label{tab:down}
  \end{table}

\subsection{Results}
As illustrated in Table \ref{tab:comp}, we adopt standard image captioning evaluation metrics to evaluate the generation performance: BLEU \cite{papineni2002bleu}, METEOR \cite{banerjee2005meteor}, and ROUGE \cite{lin2004rouge}. We also adopt $Fact_{ent}$ \cite{miura2020improving} to measure the completeness and consistency of biological entities in the generated reports.
We can observe that LSTM decoders perform relatively worse than transformer-based models.  Mem-Transformer incorporates the relational memory to implicitly model similar patterns in different reports. However, it does not show a better performance than Vanilla Transformer. The potential reason might be that pathology reports in PathText are much longer than other medical image captioning datasets, thus having too heterogeneous structures to be memorized. In our model, we adopt hierarchical position-aware modules to capture spatial semantics in the WSI. This mechanism facilitates our model to achieve the best performance no matter what the visual extractor is. 

Regarding visual extractors, ResNet and ViT do not show a large difference when pre-trained on ImageNet. But ViT pre-trained with HIPT demonstrates a significant improvement. It is not surprising because domain-aligned pre-training usually outperforms out-of-domain ImageNet pre-training in pathology \cite{kang2023benchmarking}.

Three samples of WSIs with their corresponding pathology report from different models are illustrated in Fig. \ref{fig:vis}. Our model generates more medical terms that are consistent with the ground-truth with less self-contradiction or grammar error.
The pathology reports from ground-truth contain spatial information like tumor size ("2.6 cm", "1.3$\times$1.3$\times$1.2 cm"). Vanilla Transformer fails to give a tumor spatial description (the first and third WSI) or generate not precise results (the second WSI), which reflects its disability to capture spatial features. Our model provides the perfectly right descriptions in the first and second row ("2.6 cm", "less than 5$\%$ of the tumor area") and partially correct results for the others ("1.2 cm", "2.1$\times$1.8$\times$1.8 cm"). The potential reason why 3-D size can not be accurately predicted may be that the WSI can only provide 2-D spatial information.  For the first WSI, though ground-truth only presents positive/negative results for hormone receptors, our model still has the tendency to generate spatial descriptions ("75$\%$ of the tumor cells", "not detected"). Although it is impossible to check their correctness, the corresponding binary classification of these spatial descriptions is exactly right. This comparison demonstrates that position-aware modules largely improve the spatial awareness of our model. More ablation studies can be found in \textbf{Supplementary Materials}.

\subsubsection{Slide-level Tasks.} To further verify the clinical performance of our model, we have compared WSI-related results (subtyping and Her2 prediction) which are included in the reports. We have observed that some generated reports contain the descriptions for carcinoma subtyping or Her2 prediction so that the corresponding result can be directly extracted from the text. Compared with multiple instance learning (MIL) approaches which are specially designed for WSI classification, we observe a more promising performance of our generative method, revealing the potential of our proposed visual-language learning method for computational pathology.

\section{Conclusion}
In this paper, we have shown the feasibility of pathology report generation. On the data end, we collected nearly 10000 WSI-text pairs by transforming PDF files in TCGA into concise and comprehensive pathology reports with the aid of LLMs. Our proposed PathText is the largest WSI-text dataset so far to the best of our knowledge and is going to be available to the public, which can promote visual-language learning in the pathology field. On the model end,  we introduce MI-Gen as a generative model for WSI-level descriptions. By benchmarking different baselines on the subset of PathText, we also reveal the superiority of our model in being aware of the spatial information among the WSIs and the promising performance on several slide-level tasks. In addition, other fields that utilize high-resolution images like remote sensing can be inspired by our work. 

\textbf{Acknowledgement.}
This study was partially supported by the National Natural Science Foundation of China (Grant no. 92270108), Zhejiang Provincial Natural Science Foundation of China (Grant no. XHD23F0201), and the Research Center for Industries of the Future (RCIF) at Westlake University.

\textbf{Disclosure of Interests.}
The authors have no competing interests to declare that are relevant to the content of this article.
\bibliographystyle{splncs04}
\bibliography{samplepaper}
\end{document}


\title{Supplementary}
\maketitle    
%


\titlerunning{Supplementary}
\begin{figure}[t]
  \centering
   \includegraphics[width=0.8\linewidth]{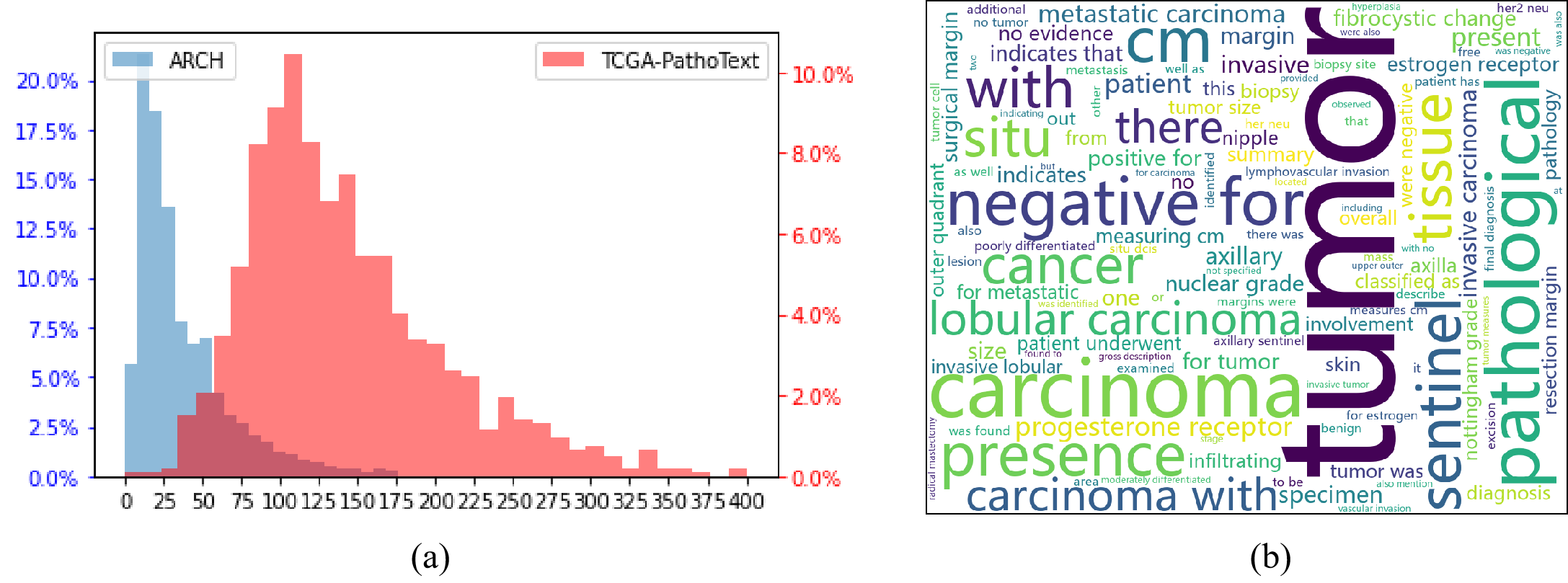}

   \caption{(a) Histogram of text lengths. It shows that TCGA-PathoText includes longer pathology reports compared to ARCH which only describes small patches. (b) Word cloud showing 100 most frequent tokens.}
   \label{fig:dataset}
\end{figure}

\begin{table}[th]
  \centering

  \resizebox{0.3\linewidth}{!}
{
    \scriptsize
  \begin{tabular}{c|c}
   \toprule
  Subset&Number of pairs\\
  \hline
  BRCA & 1061\\
  KIRC &513\\
  THCA &506\\
  UCEC &504\\
  LUSC &478\\
  LUAD &476\\
  LGG&466\\
  HNSC&450\\
  COAD&450\\
  SKCM&431\\
  STAD&416\\
  PRAD&403\\
  BLCA&386\\
  LIHC&365\\
  CESC&269\\
  SARC&247\\
  PCPG&175\\
  ESCA&156\\
  TGCT&144\\
  THYM&121\\
  OV&106\\
  KICH&94\\
  UVM&80\\
  MESO&75\\
  UCS&57\\
  ACC&56\\
  DLBC&44\\
  CHOL&39\\

    \hline
  \end{tabular}}

  \caption{Composition of TCGA-PathoText}
  \label{tab:subsets}
\end{table}

\begin{figure}[t]
  \centering
   \includegraphics[width=0.5\linewidth]{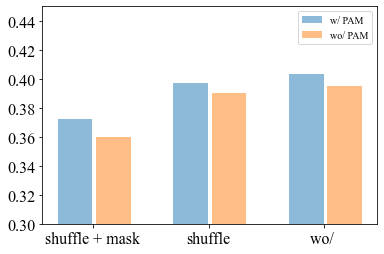}

   \caption{Effect of patch shuffling. We add noise into the spatial context by shuffling the patches in the training stage. And the test data is not shuffled or masked. We still use BLEU-1 to measure the generation performance. "shuffle + mask" represents that we shuffle and mask the tokens at the same time.}
   \label{fig:shuffle}
\end{figure}

\begin{table}[t]

  \begin{center}  
  \resizebox{0.5\linewidth}{!}{
  \begin{tabular}{c|cc}
  

   \hline
     & BLEU-4 & BLEU-1  \\
  \hline
   Single Layer & 0.092&0.377 \\
   2d sin-cos &   0.093  &0.380\\
   3$\times$3+5$\times$5+7$\times$7&0.096&0.385\\
    3$\times$3 &0.095&0.381\\
7$\times$7 &0.090&0.383\\
w/o &0.089&0.395\\
  \hline

Ours& \textbf{0.117} & \textbf{0.403} \\
 \hline
  \end{tabular}   }

  \end{center}   
    \caption{
    Effect of Position-aware Module with different structures. 'Single Layer' represents we only insert the PAM after the final encoder block.  Compared with the model without any position-aware modules, we can see that PAM can consistently improve the performance whatever the structure is. And applying convolutional layers with only one type of kernel improves the diagnosis not as well as the combination of different kernels.}
    \label{tab:ppeg}
  \end{table}